%% file: main.tex
\documentclass[10pt,twocolumn,letterpaper]{article}
\usepackage{multirow}
\usepackage{booktabs}
\usepackage{threeparttable} % 如果需要添加表注`
\usepackage{caption}
\usepackage{graphics}
\usepackage{soul}
\usepackage{wrapfig,lipsum,booktabs}
\usepackage{tablefootnote}
\usepackage{lscape} % 如果表格太宽，可以考虑横向页面
\usepackage[pagenumbers]{iccv} % To force page numbers, e.g. for an arXiv version

\input{preamble}

\definecolor{iccvblue}{rgb}{0.21,0.49,0.74}
\usepackage[pagebackref,breaklinks,colorlinks,allcolors=iccvblue]{hyperref}
\usepackage{pifont}
\newcommand{\greencheck}{\textcolor{green}{\ding{51}}}
\newcommand{\redcross}{\textcolor{red}{\ding{55}}}
\definecolor{eventgreen}{HTML}{ADC490}
\definecolor{mentalblue}{HTML}{8FC1C6}
\definecolor{causalbrown}{HTML}{EEC591}

\newcolumntype{G}{>{\columncolor{gray!40}}c}
\newcolumntype{E}{>{\columncolor{eventgreen}}c}
\newcolumntype{M}{>{\columncolor{mentalblue}}c}
\newcolumntype{C}{>{\columncolor{causalbrown}}c}

\def\confName{ICCV}
\def\confYear{2025}

\title{R\textsuperscript{3}\xspace-VQA: ``Read the Room'' by Video Social Reasoning}

\author{
    \begin{tabular}{c c c c c}
        \small\bf Lixing Niu$^{1,3,}$\thanks{Lixing Niu and Jiapeng Li contributed equally. \\
        } & \small\bf Jiapeng Li$^{2,\star}$ & \small\bf Xingping Yu$^{4}$ & \small\bf Shu Wang$^{6}$  & \small\bf Ruining Feng$^{5}$
    \end{tabular}
    \\
    \begin{tabular}{c c c c}
        \small\bf
Bo Wu$^{7}$ & \small\bf Ping Wei$^{2}$ & \small\bf Yisen Wang$^{1}$ & \small\bf Lifeng Fan$^{3}$
    \end{tabular} \vspace{3pt} \\
    \footnotesize $^1$ School of Intelligence Science and Technology, Peking University \quad{}
    \footnotesize $^2$ College of Artificial Intelligence, Xi’an Jiaotong University \\
    \footnotesize $^3$ State Key Laboratory of General Artificial Intelligence, Beijing Institute for General Artificial Intelligence \\
    \footnotesize $^4$ Yuanpei College, Peking University \quad{}
    \footnotesize $^5$ Tsinghua University\quad{} 
    \footnotesize $^6$ University of California, Los Angeles\quad{}
    \footnotesize $^7$ MIT-IBM Watson AI Lab\quad{} 
}

\begin{document}
\maketitle
\begin{strip}
\centering
\vspace{-2cm}
\includegraphics[width=\textwidth]{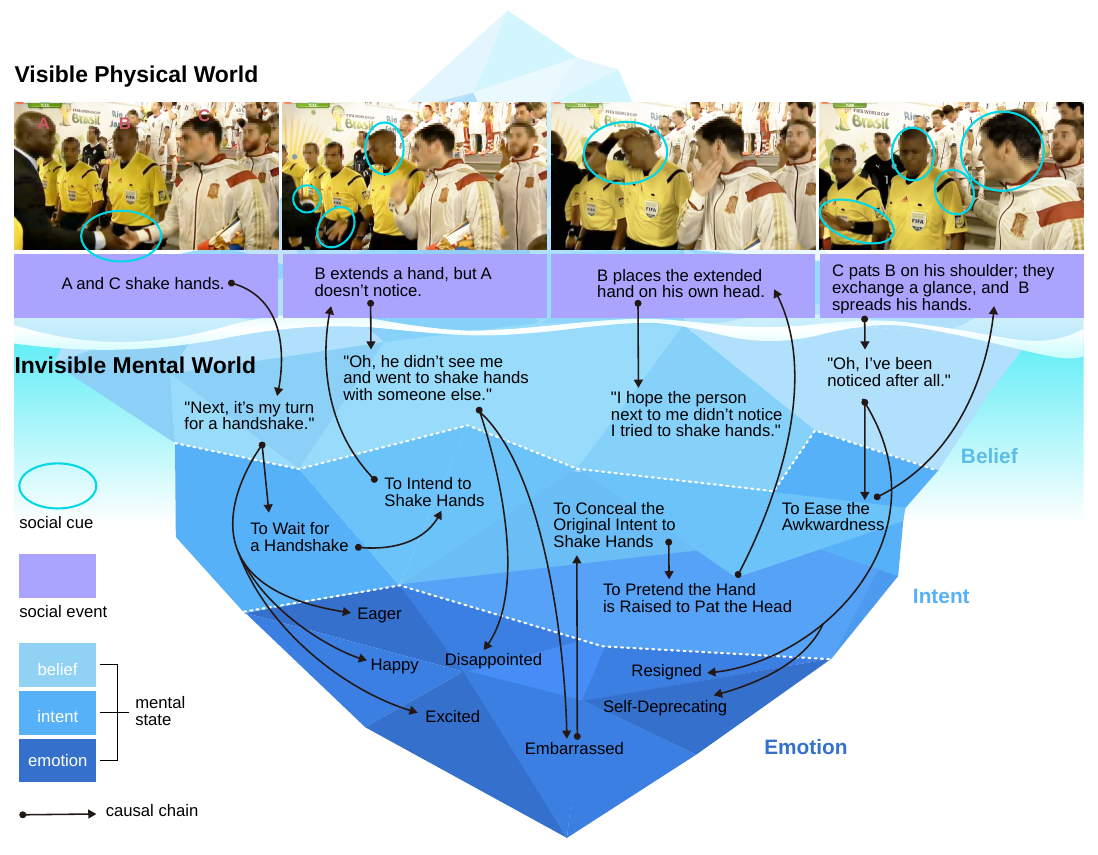}
\captionof{figure}{
The visible physical world we live in is just the tip of the iceberg compared to the vast, invisible mental world behind it \cite{zhu2020dark}. In this example\cite{youtube_video}, we can observe that a brief moment of social interaction involves a series of complex and dynamic mental activities: B extends his hand to shake with A, but A fails to notice. B then pretends that his outstretched hand was meant to touch his head, attempting to conceal his embarrassment. Despite this, C sees through B's mental state and pats him on the shoulder to offer comfort. In response, B shrugs and gestures self-deprecatingly to ease the awkwardness. Social reasoning is a critical aspect of social intelligence. However, in long-term, highly random and dynamic social interactions, capturing subtle cues, recognizing social events, accurately estimating various mental states, and identifying complex reasoning chains become progressively challenging, making social reasoning even more intricate.
}
\label{fig:teaser}
\end{strip}

\begin{abstract}
``Read the room'' is a significant social reasoning capability in human daily life. Humans can infer others' mental states from subtle social cues. Previous social reasoning tasks and datasets lack complexity (e.g., simple scenes, basic interactions, incomplete mental state variables, single-step reasoning, etc.) and fall far short of the challenges present in real-life social interactions. In this paper, we contribute a valuable, high-quality, and comprehensive video dataset named \ac{our-dataset} with precise and fine-grained annotations of social events and mental states (i.e., belief, intent, desire, and emotion) as well as corresponding social causal chains in complex social scenarios. Moreover, we include human-annotated and model-generated QAs. Our task \ac{our-dataset} includes three aspects: Social Event Understanding, Mental State Estimation, and Social Causal Reasoning. As a benchmark, we comprehensively evaluate the social reasoning capabilities and consistencies of current state-of-the-art large vision-language models (LVLMs). Comprehensive experiments show that (i) LVLMs are still far from human-level consistent social reasoning in complex social scenarios; (ii) Theory of Mind (ToM) prompting can help LVLMs perform better on social reasoning tasks. We provide some of our dataset and codes in supplementary material and will release our full dataset and codes upon acceptance.
\end{abstract}

\section{Introduction}
\label{sec:intro}
``Read the room'' requires employing Theory of Mind \cite{premack1978tom} to read others’ minds and perform social reasoning with subtle cues; it represents higher-level social intelligence, and plays a key role in helping people navigate social scenarios smoothly. Humans are innate with the ability to perceive huge hidden information from very simple cues \cite{heider1944experimental, fan2022asi_view, zhu2020dark}; however, it remains a great challenge for current AI. As illustrated in \cref{fig:teaser}, the visible physical world is just the tip of the iceberg compared to the underlying invisible mental world. In long-term and dynamic social interactions, 1) detecting subtle cues and social events, 2) estimating diverse mental states, and 3) uncovering causal interactions among variables in the physical-n-mental world, pose increasing challenges for social reasoning.

The past few years have seen large language models (LLMs) exhibit high aptitude in tasks that require reasoning capability \cite{brown2020language,wei2022emergent, kojima2022large,bubeck2023sparks, wang2024llm}. However, language models show significant limitations in reasoning for complicated tasks, such as playing long-term games or solving challenging scientific problems\cite{srivastava2022beyond, wang2024explore,mirzadeh2024gsm,glazer2024frontiermath}. Social reasoning, as an important part of complex reasoning tasks, has also proved to be hard for LLMs \cite{shapira2023cleverhans, he2023hi-tom, gu2024simpletom, wang2024evaluating}. Furthermore, it's far from enough to do social reasoning only using language. Signals from modalities beyond language are also indispensable, as they enable the inference of more complex, nuanced, and even concealed inner thoughts that individuals may not intend to reveal.
Thanks to the development of LLM, large vision-language models (LVLMs) \cite{liu2024llava, zhang2023videollama, lin2023videollava, team2023gemini1.0, team2024gemini1.5, hurst2024gpt-4o}, which can process multi-modal stimuli, have developed rapidly recently. However, there is still a lack of high-quality and comprehensive benchmarks to evaluate LVLM's (i) estimation of multiple mental states; and (ii) social reasoning accuracy and consistency in complicated social interactions.
Therefore, we craft a video question answering (VideoQA) dataset named \ac{our-dataset} with complex social interactions and fine-grained annotations of (i) social events, (ii) mental states and their transitions during social interactions, (iii) multi-step social causal chains. In addition, we include both human-annotated QAs and model-generated QAs. We evaluate the state-of-the-art (SOTA) LVLMs on \ac{our-dataset} and comprehensive experiments show that: (1) there is still a large gap for current SOTA LVLMs to improve performance; (2) LVLMs can learn from the heuristic ToM prompts and perform better on social reasoning tasks.

In summary, our contributions are three-fold: (1) We collect a novel and valuable dataset \ac{our-dataset} with complete and fine-grained annotations; (2) Comprehensive experiments show that SOTA LVLMs cannot achieve human-level performance in \ac{our-dataset} task; (3) We also find that heuristic ToM prompting helps LVLMs in our social reasoning task.

\begin{table*}[t!]
\centering
\resizebox{\linewidth}{!}{
\begin{tabular}  {c|c|ccc|cccc|cc}
        \toprule
        \multirow{2}{*}{Datasets} &  \multirow{2}{*}{Real-World} & \multicolumn{3}{c|}{Social Cue }&\multicolumn{4}{c|}{Mental State}&  \multicolumn{2}{c}{Social Reasoning}\\
        \cline{3-9}
        \cline{10-11}
  &  &  Vision & Text & Audio&   Belief&Intent& Desire&Emotion&Causality& CC\\
         \midrule
 Next-QA \cite{xiao2021nextqa}& \greencheck & \greencheck & \greencheck  & \greencheck & \redcross  &\redcross  & \redcross  &\redcross  & \redcross & \redcross \\
 ActivityNet-QA \cite{yu2019activitynetqa}& \greencheck & \greencheck & \greencheck & \greencheck & \redcross  &\redcross  & \redcross  &\redcross  & \redcross & \redcross \\
 EgoSchema \cite{mangalam2023egoschema}& \greencheck & \greencheck & \greencheck & \greencheck & \redcross  &\redcross  & \redcross  &\redcross  & \redcross & \redcross \\
 MVBench \cite{li2024mvbench}& \greencheck & \greencheck & \greencheck & \greencheck & \redcross  &\redcross  & \redcross  &\redcross & \redcross & \redcross \\
 MMBench-Video \cite{fang2024mmbench}& \greencheck & \greencheck & \greencheck & \greencheck & \redcross  &\redcross  & \redcross  &\greencheck &\redcross & \redcross \\
 Video-MME \cite{fu2024videomme}& \greencheck & \greencheck & \greencheck & \greencheck  & \redcross  &\redcross  & \redcross  &\redcross &\redcross & \redcross \\
 \midrule

MMToM-QA \cite{jin2024mmtomqa}& \redcross & \greencheck &\redcross  & \redcross &\greencheck  &\greencheck  & \redcross  &\redcross & \redcross & \redcross \\
MELD \cite{poria2018meld}& \greencheck & \greencheck & \greencheck  & \greencheck & \redcross & \redcross & \redcross & \greencheck & \redcross  & \redcross \\
BoLD \cite{luo2020bold}& \greencheck & \greencheck & \redcross  & \redcross &  \redcross & \redcross & \redcross & \greencheck & \redcross & \redcross \\
IntentQA \cite{li2023intentqa}& \greencheck & \greencheck &\greencheck & \greencheck & \redcross  &\greencheck  & \redcross  &\redcross &\redcross & \redcross \\
 \midrule
Causal-VidQA \cite{li2022causal-vidqa}& \greencheck & \greencheck & \greencheck & \greencheck & \redcross  & \redcross & \redcross & \redcross &  \greencheck & \redcross \\
CausalChaos \cite{lam2024causalchaos}& \redcross & \greencheck &\redcross  & \redcross & \redcross  &\greencheck  & \redcross  &\redcross  & \greencheck & \greencheck \\
SocialIQ \cite{zadeh2019socialiq} \& Social-IQ 2.0 \cite{social-iq-2.0} & \greencheck & \greencheck &\greencheck & \greencheck & \greencheck & \greencheck & \greencheck & \greencheck & \greencheck& \redcross \\
\bottomrule
\toprule

\textbf{R\textsuperscript{3}-VQA (Ours)}& \greencheck & \greencheck &\greencheck & \greencheck &\greencheck  &\greencheck  & \greencheck  &\greencheck & \greencheck &  \greencheck  \\
\bottomrule
\end{tabular}
}
    \caption{Dataset comparison. \textit{CC} means \textit{Causal Chain}.}
    \label{tab:dataset_comparision}
\end{table*}

\section{Related Work}
\label{sec:related_work}

\subsection{Social Reasoning}

Social reasoning is significant in achieving artificial social intelligence (ASI) \cite{fan2022asi_view}. With ToM as a core component, social reasoning requires the ability to identify mental states (such as intentions and emotions) and to interpret behavior. Today, LLMs have shown some social reasoning capabilities, but most of them remain limited to textual domain \cite{sap2019socialiqa, shapira2023cleverhans, kim2023fantom, xu2024opentom, ullman2023llmfailtrivialtom}. While text is essential for social reasoning, it cannot encompass all forms of information, such as facial expressions, tone of voice, and other key social cues like gaze, etc. These often reveal important insights that individuals may wish to conceal in dialogue but unintentionally convey through nonverbal cues. Such information is crucial for understanding the inner thoughts and emotions of others. Therefore, it is vital to develop more challenging social reasoning tasks that incorporate multimodal inputs. Some works estimate beliefs and intentions \cite{puig2020watchandhelp, puig2023nopa}, but the tasks are quite simple, such as inferring the object's location and which object others want to obtain. To our best knowledge, we are the first VideoQA dataset that focuses on challenging social reasoning in real-life world, with fine-grained and complete annotations of multi-modality social cues, all kinds of mental states and causal chains.

\subsection{Video Understanding}
Video understanding tasks are diverse, but can generally be divided into two categories: one category focuses on low-level, fact-based understanding tasks, such as object tracking \cite{VOT_TPAMI, huang2019got_10k}, action recognition \cite{caba2015activitynet, soomro2012ucf101}, and active speaker detection \cite{chaudhuri2018ava_active_speaker}, etc.; the other category aims on high-level tasks that require extensive video understanding, such as video reasoning \cite{2021star}, video captioning \cite{krishna2017ava_dense_caption, zhou2018youcook2, wang2019vatex} and video question answering \cite{xiao2021nextqa, yu2019activitynetqa}, etc.
We compare the \ac{our-dataset} with other VideoQA datasets in \cref{tab:dataset_comparision}. 
Next-QA \cite{xiao2021nextqa}, ActivityNet-QA \cite{yu2019activitynetqa}, EgoSchema \cite{mangalam2023egoschema}, MVBench \cite{li2024mvbench}, MMBench-Video \cite{fang2024mmbench} and Video-MME \cite{fu2024videomme} focus on factual understanding or causal reasoning, ignoring the understanding and causal reasoning of mental states. Although the videos include audio, they focus more on visual cues. MMToM-QA \cite{jin2024mmtomqa}, MELD \cite{poria2018meld}, BoLD \cite{luo2020bold}, and IntentQA \cite{li2023intentqa} focus on simple mental states and lack social causal reasoning. \citet{li2022causal-vidqa} only focus on the causal relationship between factual events and lack of attention to causality in social interactions. \citet{lam2024causalchaos} involves causal reasoning of mental states; however, its video source is cartoons, which differ significantly from real-life scenarios. Moreover, it does not incorporate language-based social cues, which are rich in information. 
\citet{zellers2019vcr} focuses on commonsense reasoning, intent understanding, and behavior predicting, but involves no more mental states; and many of its questions heavily rely on commonsense to answer rather than the context. \citet{zadeh2019socialiq, social-iq-2.0} covers multiple aspects of social intelligence but lacks examination of capturing and reasoning about visual cues, as well as fine-grained causal chains. 

\section{The \ac{our-dataset} Challenge}
\label{sec:r3-vqa}

We design our dataset in a natural, intuitive and reasonable way, based on the foundational Theory of Mind \cite{premack1978tom}, the belief-desire-intention (BDI) framework \cite{bratman1987intention}, and Bandura's social cognitive theory, particularly the concept of triadic reciprocal determinism \cite{bandura1986social, bandura1989human}, as well as other modern studies on social cognition \cite{tomasello2010origins, pearl2014probabilistic, pearl2009causality, reisenzein2006emotions, reisenzein2009emotional, puica2013emotional, schlaffke2015shared, fan2022asi_view}. Our dataset systematically incorporates both the physical and mental dimensions of social interactions. We include key mental state variables (belief, intention, desire, and emotion) alongside observable physical variables (social events encompassing action, expression, dialogue, and other social cues). We also include comprehensive and dynamic causal interactions among these variables, capturing the complex interplay between internal states and external behaviors and environments. This comprehensive framework positions our dataset as a valuable testbed for developing and evaluating computational models of social reasoning.

\begin{figure}[t!]
    \centering
    \includegraphics[width=\linewidth]{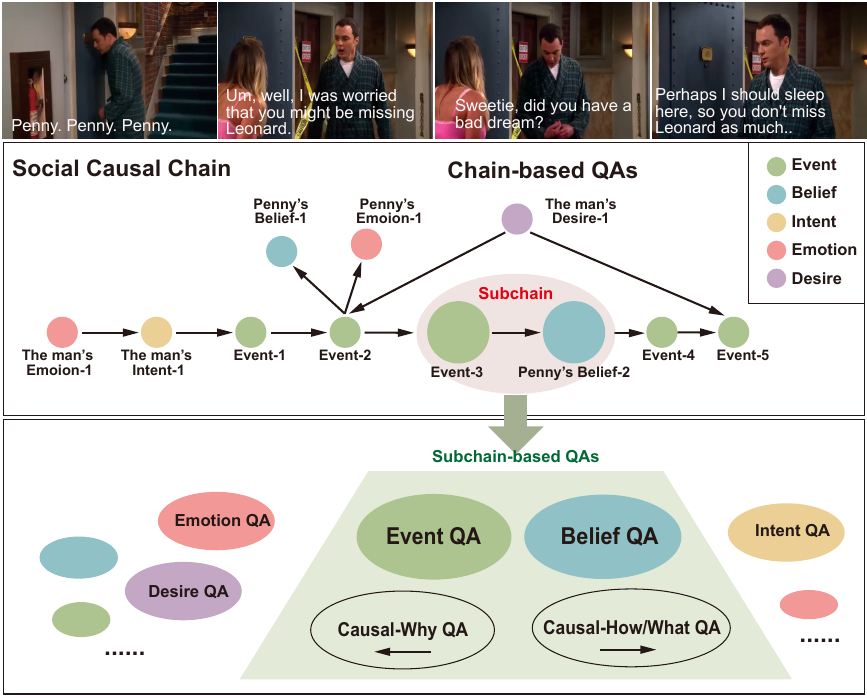}
    \caption{We generally illustrate our dataset design (see \cref{subsec: dataset design}).}
    \label{fig:data_overview}
\end{figure}

\subsection{Dataset Design}
\label{subsec: dataset design}
\ac{our-dataset} dataset differs from 
traditional VideoQA datasets in the following aspects: (i) We annotate complete and fine-grained social causal chains; (ii) Apart from human annotated QAs, we also generate QAs based on the social causal chains; (iii) We support comprehensive evaluation of many aspects of social reasoning capabilities, as well as reasoning consistency with our social causal chains.
We illustrate the design and structure of our dataset via an example in \cref{fig:data_overview}. 

Each video has one or more social causal chains. We denote a social causal chain as $g \in \mathcal{G}$, and a subchain as $g^{sub} \in \mathcal{G}^{sub}$. Here, $\mathcal{G}$ and $\mathcal{G}^{sub}$ respectively represent all the social causal chains and subchains in \ac{our-dataset}. Note that a chain consists of multiple subchains: $g = \{g^{sub}\}$. A subchain $g^{sub}$ comprises one result node $v^{1}$ and one or several reason nodes $\{v^{0}_i\}$. All reason nodes point to the result node via causal edges $\{e_i\}$. Every reason node $v^0_i$ is necessary to deduce the result node $v^1$.

For each social causal chain $g$, we generate a set of related QAs according to the following rules:

\begin{itemize}
    \item For each node $v \in g$, we generate a \textit{Event Understanding} or \textit{Mental State Estimation} QA, depending on its node content. The content of a node is either about an event or a mental state (i.e., belief, intent, emotion, or desire).
    \item For each subchain $g^{sub}$, we generate a \textit{Causal-Why} QA and a \textit{Causal-How/What} QA, depending on whether the reasoning for the subchain is abductive or deductive. 
\end{itemize}

\begin{figure}[t!]
    \centering
    \includegraphics[width=\linewidth]{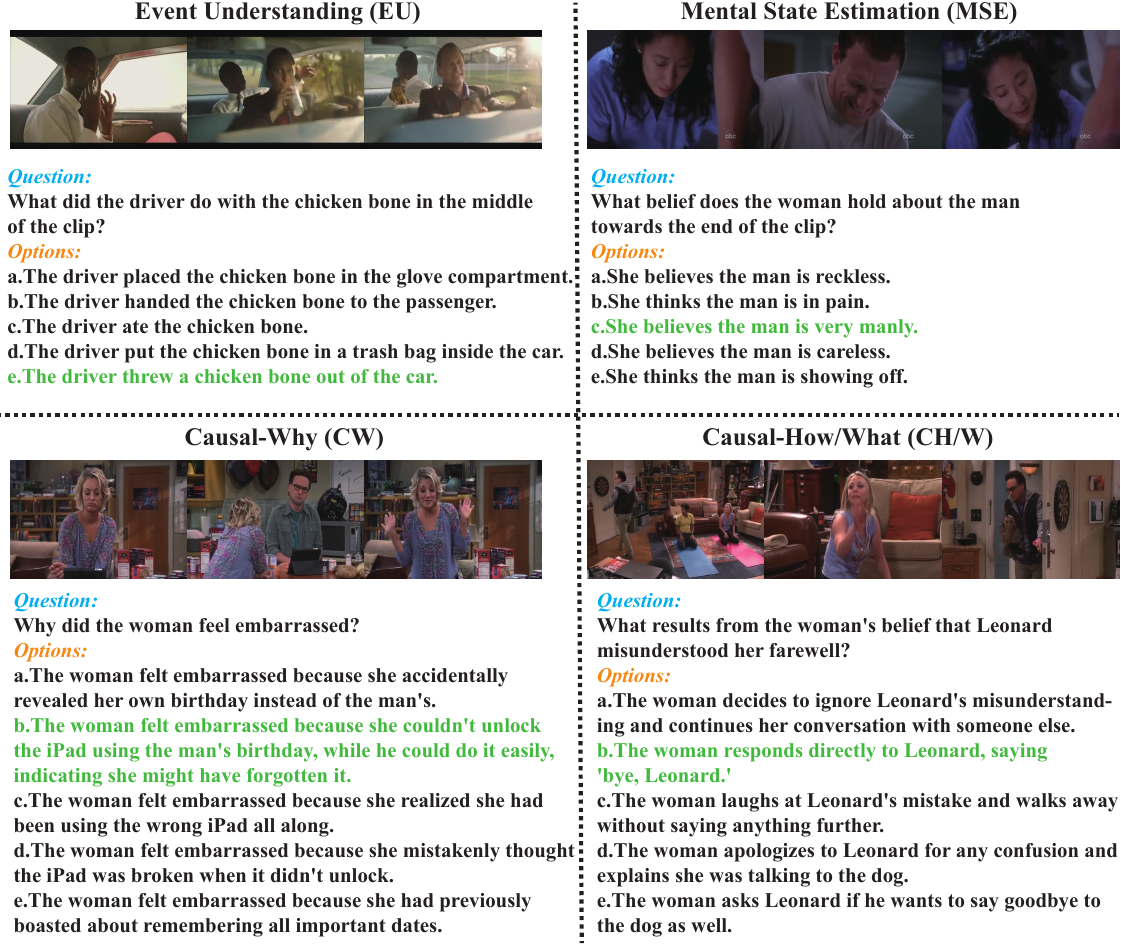}
    \caption{Examples of each QA type. The option marked in \textcolor{green}{green} is the correct answer.}
    \label{fig:qa_types}
\end{figure}

Therefore, we have four types of QAs in total. We provide four examples selected from our dataset for the four QA types respectively in \cref{fig:qa_types}: 

\textbf{Event Understanding (EU).} Event understanding is the premise of social reasoning. We generate a factual QA for each event node. To ensure QA diversity, we vary the ways we frame questions, such as by locating events through (i) the period of the event, as in ``What happens at the end of the clip?'', or (ii) specific parts of the event, as in ``What does Person B do when Person A reaches out her hands?''. 

\textbf{Mental State Estimation (MSE).} Mental state estimation is a crucial aspect of social intelligence. We consider typical mental states including belief, intent, desire, and emotion. We generate an inferential QA for each mental state node, such as ``How does someone feel at the end of the clip?'' 

\textbf{Causal-Why (CW) \& Causal-How/What (CH/W).} 
\textit{Causal-Why} QAs focus on abductive reasoning and ask about reasons for a result, such as the reasons why \textit{the woman feels embarrassed} in the third example of \cref{fig:qa_types}. On the contrary, \textit{Causal-How/What} QAs focus on deductive reasoning and ask about the effect of the cause, such as the result of \textit{the woman's belief} in the fourth example of \cref{fig:qa_types}. Note that we use variations such as ``How'' and ``What'' to enhance the diversity of question formats. One example of \textit{Causal-How} QA is: \textit{Question}: ``How did Sheldon's revelation affect Penny's emotions?'' \textit{Answer}: ``Sheldon's comment about the ring's true value led to Penny's disappointment.''Please note that, although we currently generate only single-step causal QAs, by leveraging fine-grained causal chains, we can also generate multi-step causal QAs, which will hugely increase the scale of our dataset and evaluate models' capabilities more comprehensively.

\begin{figure*}
    \centering
    \includegraphics[width=\linewidth]{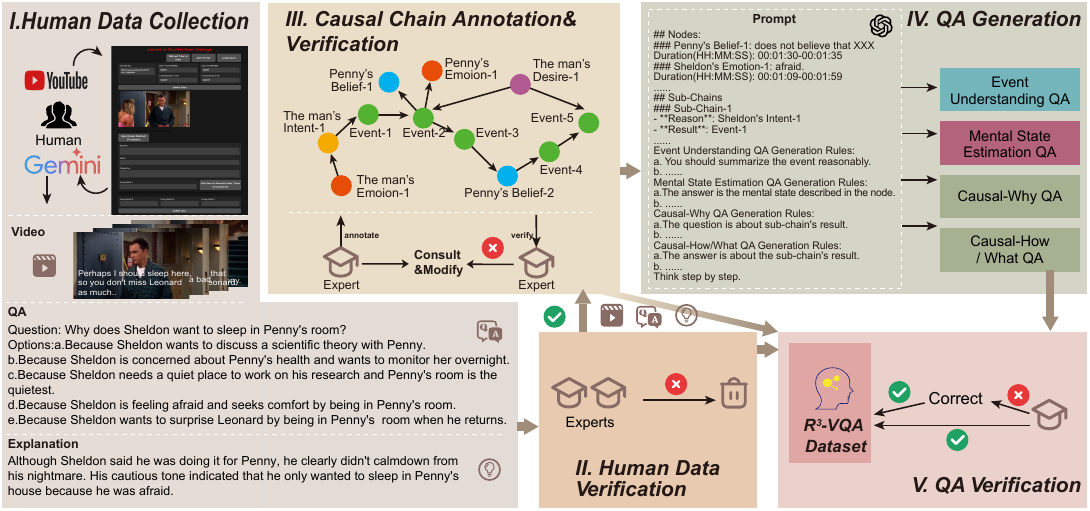}
    \caption{Our dataset construction pipeline, which consists of five stages.} 
    \label{fig:data-pipeline}
\end{figure*}

\subsection{Dataset Construction and Annotation}
\label{subsec: data construction}
\paragraph{Principles and Standards.} 
We establish principles and standards for collecting and annotating video data. 
Each data sample includes a video clip, a question, an answer, four incorrect options, and an explanation, focusing on social interactions and human mental states.
Questions should emphasize causal relationships between events and mental states and involve at least one type of mental state. The answer must be inferred from the social cues present in the video clip, without reliance on external background information. 
Explanations detail the underlying reasoning process and causal relationship.
Characters are referenced by descriptive phrases, or by their names if mentioned in the clip. Additionally, we prioritize video clips that are under three minutes and in English.

Given that some existing LVLMs have already been trained on extensive video datasets, we need to avoid data that may overlap with their training sets. In a preliminary experiment, we find that some LVLMs such as Gemini 1.5 Pro \cite{team2024gemini1.5} is trained with a wide range of popular movies and TV shows. This is not fair for the evaluation. Thus, to avoid data pollution and to make the dataset more challenging, we use Gemini 1.5 Pro as a touchstone to identify and retain videos and QA pairs that it answers incorrectly.
\textbf{Human Data Collection.} We build an online data collection platform
using Gradio\footnote{\url{https://www.gradio.app/}}. To collect data, we organize the 'Read the Room Challenge', recruiting 44 educated adult participants from social media and 157 from Prolific\footnote{\url{https://www.prolific.com/}}. All participants first complete tutorials and a brief quiz covering our requirements and instructions for using our online platform. Then the participants search for videos on YouTube and design corresponding QA pairs to challenge Gemini 1.5 Pro on our platform. A challenge is considered successful if Gemini 1.5 Pro answers incorrectly, and such video and QA pairs are retained for further verification. Participants are also required to provide detailed explanations to support subsequent causal chain annotations. This process yields a total of 667 data samples, each with a video clip, a QA, and an explanation.

\textbf{Human Data Verification.} 
Although we receive many videos and QA pairs that stump Gemini 1.5 Pro, some issues arise in the collected data. For instance, some questions have more than one plausible answer among the options. Additionally, certain video clips are unrelated to ``read the room'' behaviors, as they lack social interactions and human mental states. Some questions require excessive background information or using character names not mentioned in the video. We engage 24 experts to verify the data collected in the previous stage and exclude any unqualified samples. Each data sample is verified independently by two experts, and only those approved by both are retained as valid data. This process helps to minimize subjectivity and bias in our dataset. After that, we obtain 316 unique video clips, 316 QAs, and 316 explanations.

\textbf{Causal Chain Annotation\&Verification} We asked the 24 experts from the previous stage to annotate causal chains for the validated data. After training them on strict principles and standards, we instructed them to follow the causal chain format described in \cref{subsec: dataset design}. Each data sample was annotated by one of the experts who verified it in the previous stage. The whole annotation process included three steps: (i) reviewing the QA pair and explanation, (ii) annotating nodes for relevant events and mental states, and (iii) linking nodes based on causal relationships to form subchains. 
Experts could translate participants’ explanations into causal chains or create new causal chains if they wanted to expand on or modify the original explanations. In annotated chains, each node is linked to the relevant character(s) and includes the corresponding start and end times within the video clip. 
After that, the other of the two experts verify whether the annotated causal chains are reasonable. If the expert deems chains unreasonable, they are revised through consultation and modification by both experts. These two experts must reach a consensus and determine the final version of causal chains.

\textbf{QA Generation.}
We used GPT-4o to generate a set of related QA pairs for each causal chain, following the QA generation rules introduced in \cref{subsec: dataset design}. Firstly, we generate \textit{Event Understanding} QAs or \textit{Mental State Estimation} QAs for all the nodes. Secondly, we generate \textit{Causal-Why} QAs and \textit{Causal-How/What} QAs for all subchains.  

\textbf{QA Verification.} We invited experts from the causal chain annotation stage to verify the QA pairs generated from their annotated causal chains, following these standards: (i) QAs adhere to the generation rules, (ii) time references in the QAs clearly and accurately indicate their relative positions within video clips, (iii) questions and answers fully encompass the content of the corresponding nodes or subchains, and (iv) each question has only one correct option among five choices. If a QA pair did not meet these standards, experts could either correct it according to our guidelines or delete it if modification proved challenging. This process resulted in 4840 generated and verified QAs. 

Finally, \ac{our-dataset} contains annotated social causal chains, human designed QAs and generated QAs.

\subsection{Dataset Statistical Analysis}
\label{subsec: data analysis}

\begin{figure}[t]
    \centering
    \begin{subfigure}{0.23\textwidth}
        \centering
        \includegraphics[width=\linewidth]{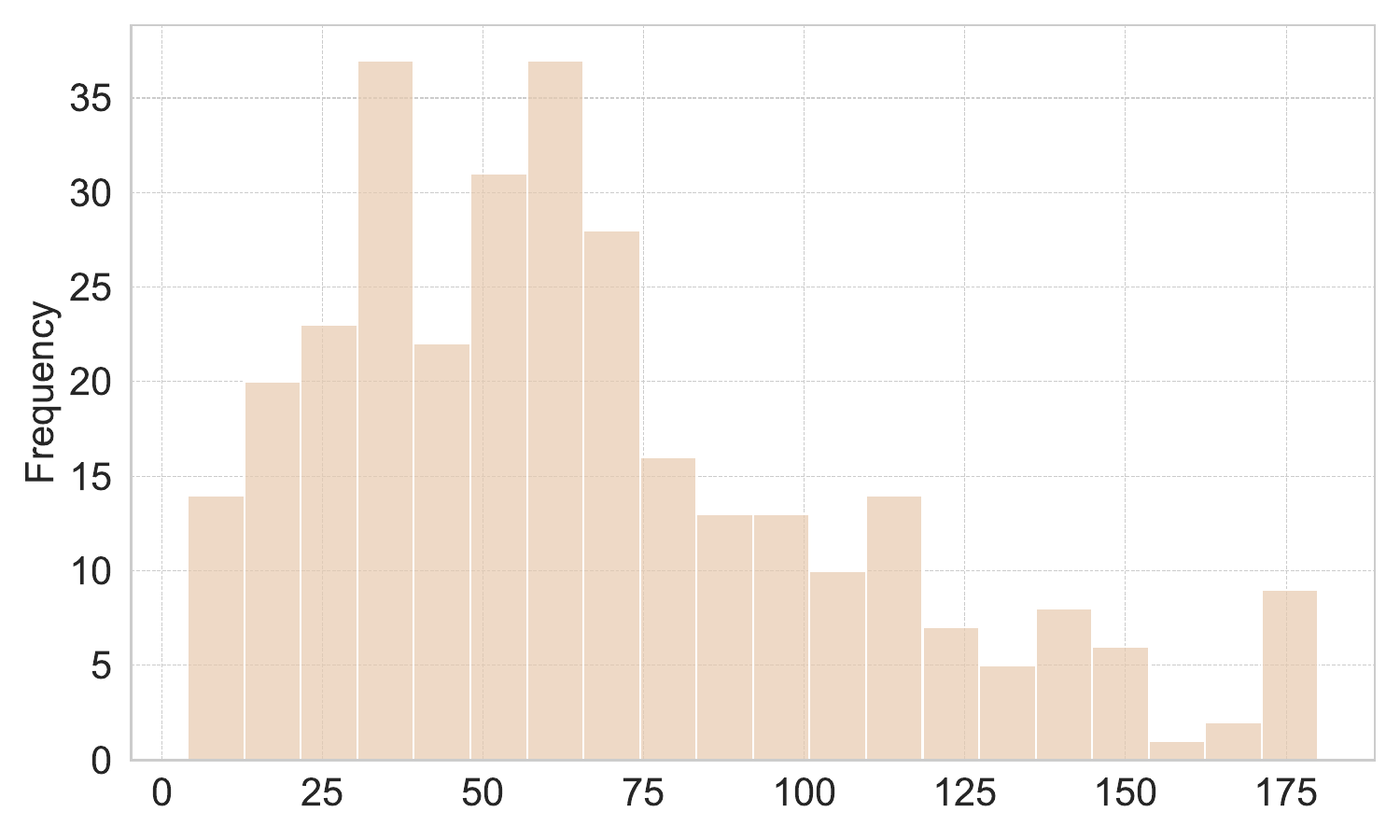}
        \caption{Video Length}
        \label{fig: video length}
    \end{subfigure}
    \hfill
    \begin{subfigure}{0.23\textwidth}
        \centering
        \includegraphics[width=\linewidth]{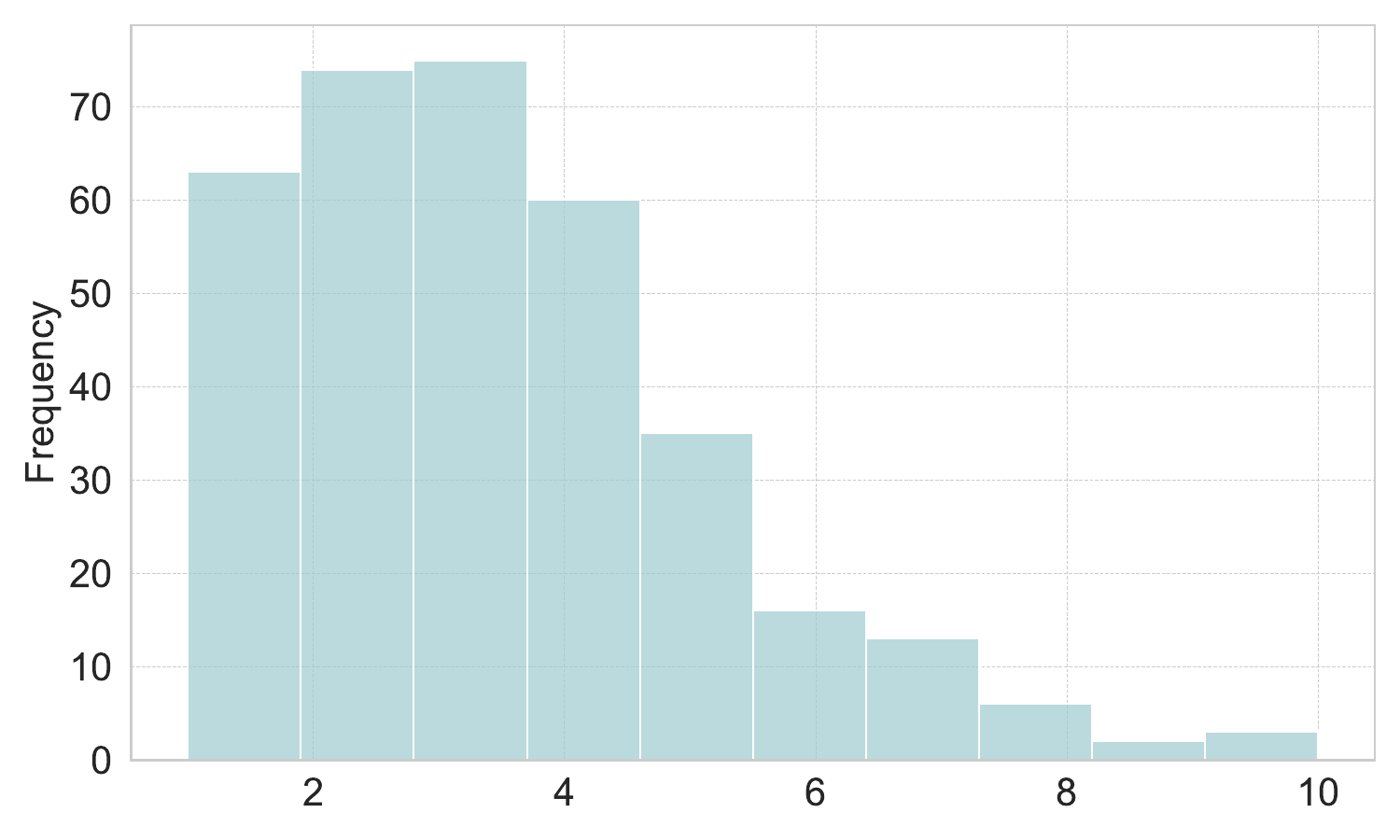}
        \caption{Causal Chain Length}
        \label{fig:chain length}
    \end{subfigure}

    \begin{subfigure}{0.23\textwidth}
        \centering
        \includegraphics[width=\linewidth]{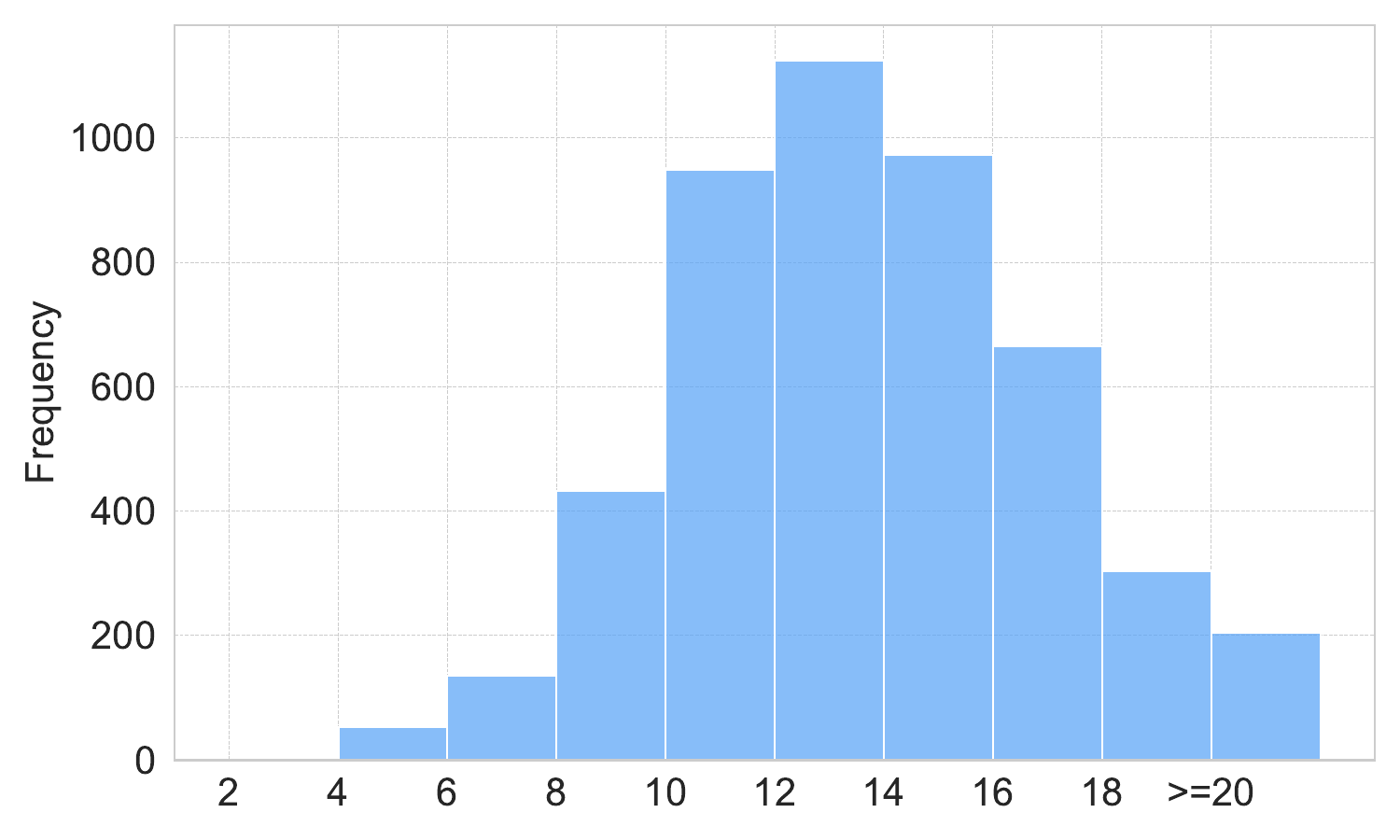}
        \caption{Question Length}
        \label{fig:question length}
    \end{subfigure}
    \hfill
    \begin{subfigure}{0.23\textwidth}
        \centering
        \includegraphics[width=\linewidth]
        {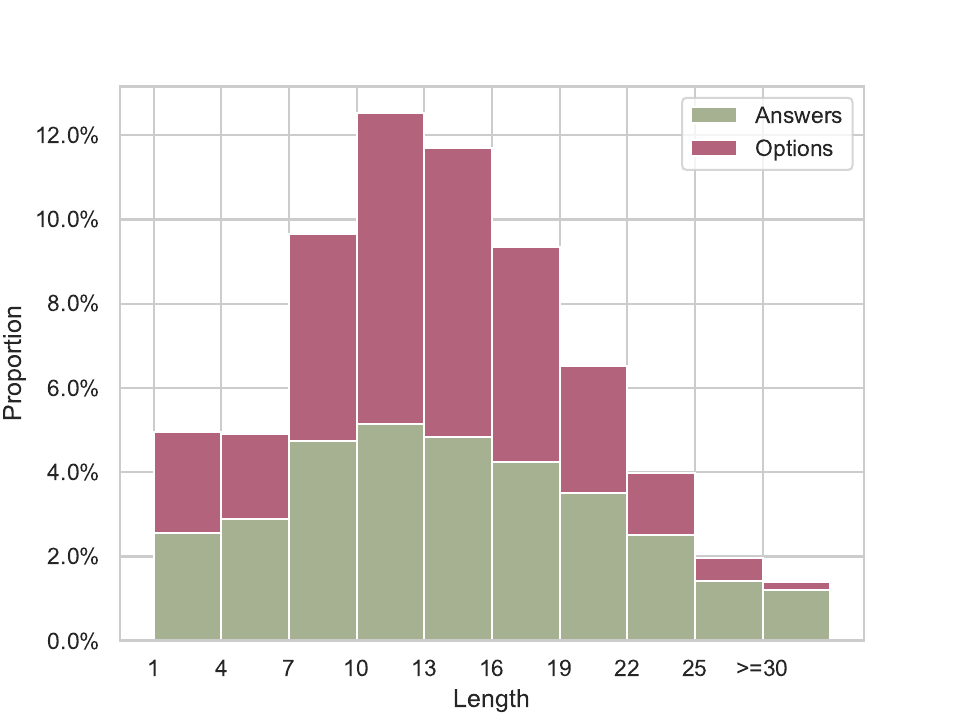}
        \caption{Answer and Option Lengths}
        \label{fig:ao lengths}
    \end{subfigure}

    \caption{Statistics of generated QA data.}
    \label{fig:data-analysis}
    % \vspace{-0.5cm}
\end{figure}

\textbf{Video Statistics.} Our \ac{our-dataset} dataset contains 316 videos, each annotated with one or more causal chains and multiple QA pairs. \cref{fig: video length} shows the distribution of video durations. All videos are under 180 seconds, with an average duration of 66.6 seconds. The increased video length amplifies the challenge of social reasoning, as identifying relevant social cues becomes more difficult, and causal chains may become longer with more steps, involving a greater variety of events and mental states and requiring tracking of all dynamic changes of all states throughout the video. 

\textbf{Causal Chain Statistics.} \cref{tab:dataset statistics} presents the statistics of causal chains in our \ac{our-dataset} dataset. The dataset includes 347 causal chains composed of 2198 nodes and 1406 single-step subchains. These nodes are categorized into 997 \textit{Event} nodes, 321 \textit{Belief} nodes, 361 \textit{Intent} nodes, 42 \textit{Desire} nodes, and 477 \textit{Emotion} nodes. Aside from \textit{Desire}, each mental state category has over 300 nodes, providing ample cases to assess specific mental states. In total, 1201 mental state nodes indicate a rich presence of mental state dynamics. The lengths of causal chains range from 1 to 10 steps, with an average length of 3.3 steps, as shown in \cref{fig:chain length}. Over 60\% have no less than three steps, suggesting that the videos require complex and in-depth reasoning.

\textbf{QA Statistics.} As detailed in \cref{tab:dataset statistics}, \ac{our-dataset} comprises 4840 generated QAs and 316 human-designed QAs. \cref{fig:question length} and \cref{fig:ao lengths} show the distributions of question, answer, and wrong option lengths among generated QAs. The average question length is 13.2 words—longer than many popular VideoQA datasets \cite{Tapaswi_2016_CVPR, zadeh2019socialiq, fu2024videomme, yu2019activitynetqa, Zeng_Chen_Chuang_Liao_Niebles_Sun_2017}. Answers and wrong options average 14.4 and 12.8 words respectively, making the distractors similar in length to the correct answers and thus more challenging. 
With a total of 5.1K QAs, our dataset is comparable to SocialIQ 1.0/2.0 (7K/6K) and CausalChaos (4.9K).

\begin{table}[t]
    \centering
    \resizebox{\linewidth}{!}{
    \begin{tabular}{c|ccccc|c}
        \toprule
         \multirow{2}{*}{\textbf{Videos}} &\multicolumn{5}{c|}{\textbf{Model Generated QAs}} & \textbf{Human Designed}\\ \cline{2-6}
    & EU& MSE& CW& CH/W& Overall& \textbf{QAs}\\
          \midrule
          316 & 997 & 1201& 1405& 1237& 4840& 316\\
        \bottomrule
    \end{tabular}
    }
    \\
    \resizebox{\linewidth}{!}{
    \begin{tabular}{c|c|cccccc}
        \toprule
          \multirow{2}{*}{\textbf{Chains}} &\multirow{2}{*}{\textbf{Subchains}}&\multicolumn{6}{c}{\textbf{Nodes}} \\ \cline{3-8}
 & &  Event& Belief&Intent&Desire& Emotion& Overall \\
          \midrule
          347 & 1406 & 997 & 321 & 361 & 42 & 477& 2198\\
        \bottomrule
    \end{tabular}   
    }
\caption{Statistics of \ac{our-dataset} dataset}
\label{tab:dataset statistics}
\end{table}

\section{Experiments}
\label{sec:exps}
Our experimental methodology consists of two main components: the \textbf{Generated QA Evaluation} and the \textbf{Human Designed QA Evaluation}. In the first component, we build a benchmark by evaluating various models, including several open-source and prominent closed-source models. In the second component, we use some of the better-performing models from the first evaluation to test on human-designed QA pairs.

% \end{wraptable}
\subsection{Configuration}
Our evaluation is based on VLMEval\footnote{\url{https://github.com/open-compass/VLMEvalKit/}} repository, an open-source toolkit for assessing LVLMs across multiple benchmarks without extensive data preparation. By extending VLMEval, we incorporates our dataset as a new benchmark.

\begin{table*}[htbp]
\centering
\small
\caption{Generated QA Evaluation Results (\%). MSE: mental state estimation. EU: event understanding. CW: causal why. CH: causal how. $Cons^c$: chain consistency. $Cons^c$: sub-chain consistency.}
\label{tab:corrected_table}
\small{
\resizebox{\textwidth}{!}{
\begin{tabular}{llMMMMMcCCcGG}
\toprule
\multirow{2}{*}{\textbf{Model}} & \multirow{2}{*}{\textbf{Setting}} & \multicolumn{5}{c}{\textit{MSE}} & \multirow{2}{*}{\textit{EU}} & \multirow{2}{*}{\cellcolor{white}{\textit{CW}}} & \multirow{2}{*}{\cellcolor{white}{\textit{CH/W}}} & \multirow{2}{*}{\textit{Overall}} & \multirow{2}{*}{\cellcolor{white}{$Cons^{c}$}} & \multirow{2}{*}{\cellcolor{white}{$Cons^{sc}$}} \\ \cmidrule(lr){3-7}
 &  & \cellcolor{white}{\textit{Emotion}} & \cellcolor{white}{\textit{Belief}} & \cellcolor{white}{\textit{Intent}} & \cellcolor{white}{\textit{Desire}} & \cellcolor{white}{\textit{Overall}} &  & \cellcolor{white} & \cellcolor{white} &  &\cellcolor{white}  &\cellcolor{white}  \\ \midrule
Random & - & 20 & 20 & 20 & 20 & 20 & 20 & 20 & 20 & 20 & - & - \\ \midrule
\multirow{3}{*}{Video-LLaVA\cite{lin2023video}} & - & 18.03 & 18.69 & 20.22 & 16.67 & 18.82 & 20.26 & 20.57 & 21.58 &20.33 & 0.00 & \textit{0.14} \\  
 & +Sub & 19.08 & 19.63 & 24.38 & 21.43 & 20.90 & 28.28 & 21.64 & 22.88 &\textbf{23.14} & 0.00 & \textbf{\textit{0.36}} \\ 
  & + Sub \& HTP &  18.87 & 19.63 & 22.16 & 21.43 & 20.15 & 25.78 & 20.64 & 22.47 &22.05 & 0.00 & \textit{0.21} \\
 \midrule

\multirow{3}{*}{Idefics2-8B\cite{laurenccon2024matters}} 
& - & 10.48 & 8.41 & 10.80 & 2.38 & 9.74 & 8.93 & 8.90 & 11.48 & 9.77 & 0.00 & \textit{0.07} \\
 & + Sub  & 21.80 & 21.50 & 22.99 & 19.05 & 21.98 & 20.56 & 22.99 & 22.64 & \textbf{22.15} & 0.29 & \textit{1.14} \\
 & + Sub \& HTP & 19.08 & 18.69 & 21.61 & 16.67 & 19.65 & 20.26 & 21.07 & 20.94 & 20.52 & 0.00 & \textit{0.92} \\
 \midrule

\multirow{3}{*}{mPLUG-Owl3\cite{ye2024mplug}} & - & 48.01 & 59.50 & 42.94 & 45.24 & 49.46 & 52.46 & 67.90 & 63.22 & 58.95 & 2.88 & \textit{13.66} \\
 & + Sub  & 51.15 & 71.03 & 50.97 & 50.00 & 56.37 & 75.03 & 74.66 & 70.82 & \textbf{69.21} & 9.22 & \textit{25.39} \\
 & + Sub \& HTP  & 23.06 & 28.97 & 27.15 & 35.71 & 26.31 & 41.62 & 42.85 & 36.05 & 36.76 & 0.58 & \textit{3.56} \\
 \midrule

\multirow{3}{*}{Phi-3.5-Vision\cite{abdin2024phi}} & - & 48.85 & 59.50 & 49.86 & 35.71 & 51.54 & 53.46 & 72.31 & 70.74 & 62.87 & 4.09 & \textit{16.36} \\
 & + Sub  & 46.75 & 62.62 & 53.46 & 40.48 & 52.79 & 68.71 & 76.16 & 72.92 & \textbf{68.00} & \textbf{8.65} & \textbf{\textit{23.47}} \\
 & + Sub \& HTP  & 48.22 & 61.06 & 51.80 & 38.10 & 52.37 & 69.21 & 74.59 & 72.76 & 67.50 & 8.07 & \textit{22.69} \\ \midrule

 \multirow{3}{*}{Idefics3-8B-Llama3\cite{laurenccon2024building}} & - & 40.88 & 57.94 & 48.75 & 35.71 & 47.63 & 51.15 & 63.70 & 62.49 & 56.82 & 3.17 & \textit{12.02} \\
 & + Sub & 45.49 & 71.65 & 55.12 & 54.76 & 55.70 & 74.32 & 74.45 & 69.44 & 68.49 & 8.07 & \textit{23.83} \\
 & + Sub \& HTP  & 42.35 & 70.09 & 56.23 & 50.00 & 54.20 & 76.43 & 74.16 & 69.68 & \textbf{68.53} & \textbf{9.22} & \textbf{\textit{25.96}} \\ \midrule

 \multirow{3}{*}{PLLaVA-7B\cite{xu2024pllava}} & - & 28.93 & 26.48 & 27.98 & 14.29 & 27.48 & 29.29 & 31.10 & 32.74 & 30.25 & \textbf{0.29} & \textbf{\textit{1.42}} \\
 & + Sub  & 22.64 & 22.43 & 26.59 & 23.81 & 23.81 & 36.21 & 29.25 & 29.02 & \textbf{29.28} & 0.00 & \textit{1.21}  \\
 & + Sub \& HTP  & 19.08 & 19.63 & 22.99 & 16.67 & 20.32 & 31.09 & 24.41 & 24.01 & 24.67 & 0.00 & \textit{0.71} \\
 \midrule 
 
\multirow{3}{*}{PLLaVA-13B\cite{xu2024pllava}} & - & 23.48 & 22.43 & 22.71 & 16.67 & 22.73 & 23.77 & 33.10 & 30.72 & 28.00 & 0.00 & \textit{0.64} \\
 & + Sub  & 27.04 & 30.84 & 26.04 & 26.19 & 27.73 & 37.61 & 38.58 & 35.57 & \textbf{34.92} & \textbf{0.58} & \textbf{\textit{2.56}} \\ 
 & + Sub \& HTP   & 23.90 & 24.92 & 23.27 & 19.05 & 23.81 & 33.30 & 34.31 & 31.45 & 30.76 & 0.29 & \textit{1.56} \\
 \midrule 

 \multirow{3}{*}{PLLaVA-34B\cite{xu2024pllava}} & - & 49.90 & 59.81 & 55.40 & 54.76 & 54.37 & 52.46 & 69.54 & 70.57 & 62.52 & 5.19 & \textit{16.22} \\
  & + Sub    & 53.67 & 70.40 & 68.14 & 69.05 & 63.03 & 78.44 & 77.30 & 78.42 & 74.28 & \textbf{14.12} & \textbf{\textit{33.50}} \\
 & + Sub \& HTP  & 53.04 & 71.03 & 71.19 & 71.43 & 63.95 & 78.03 & 79.00 & 78.33 & \textbf{74.90} & 13.83 & \textbf{\textit{33.50}} \\

 \midrule 

\multirow{3}{*}{InternVL2-8B\cite{chen2023internvl}} & - & 47.17 & 51.09 & 42.11 & 47.62 & 46.71 & 46.94 & 61.78 & 58.69 & 54.19 & 3.17 & \textit{11.17} \\
 & + Sub & 49.90 & 68.54 & 54.57 & 52.38 & 56.37 & 73.22 & 73.38 & 68.31 & 67.83 & \textbf{8.06} & \textit{25.04} \\
 & + Sub \& HTP  & 50.10 & 69.16 & 55.12 & 54.76 & 56.87 & 72.12 & 73.81 & 69.93 & \textbf{68.26} & 7.78 & \textbf{\textit{26.17}} \\ \midrule

\multirow{4}{*}{InternVL2-26B\cite{chen2023internvl}} & - & 42.14 & 49.22 & 45.43 & 45.24 & 45.13 & 47.44 & 59.86 & 57.56 & 53.06 & 3.17 & \textit{10.60} \\
& Only Sub & 45.49 & 70.09 & 54.85 & 54.76 & 55.20 & 69.71 & 72.88 & 69.44 & 66.96 & 9.51 & \textit{23.83} \\
 & + Sub  & 46.12 & 71.96 & 58.73 & 57.14 & 57.20 & 73.62 & 74.45 & 71.22 & \textbf{69.17} & \textbf{13.26} & \textbf{\textit{27.03}} \\
 & + Sub \& HTP & 46.75 & 69.47 & 57.89 & 57.14 & 56.54 & 72.42 & 74.38 & 71.38 & 68.78 & 10.66 & \textit{25.68} \\ \midrule

\multirow{3}{*}{InternVL2-76B\cite{chen2023internvl}} & - & 42.35 & 63.86 & 48.48 & 45.24 & 50.04 & 57.87 & 69.11 & 66.61 & 61.43 & 5.75 & \textit{15.58} \\
 & + Sub & 53.46 & 75.70 & 65.10 & 64.29 & 63.28 & 81.34 & 79.93 & 76.39 & \textbf{75.19} & 17.29 & \textbf{\textit{35.99}} \\
 & + Sub \& HTP  & 54.09 & 73.52 & 63.71 & 69.05 & 62.70 & 81.75 & 78.86 & 76.56 & 74.86 & \textbf{18.16} & \textit{35.14} \\ 
 \midrule 

\multirow{4}{*}{GPT-4o Mini\tablefootnote{\url{https://openai.com/index/gpt-4o-mini-advancing-cost-efficient-intelligence/}, 2024-07-18}} & - & 43.19 & 57.63 & 55.40 & 52.38 & 51.04 & 62.39 & 73.52 & 69.36 & 64.59 & 6.05 & \textit{18.92} \\
 & + Sub  & 43.19 & 57.63 & 55.40 & 52.38 & 51.04 & 62.39 & 73.52 & 69.52 & 64.63 & 6.05 & \textit{18.99} \\
 & + Sub \& HTP  & 49.27 & 73.52 & 64.27 & 71.43 & 61.03 & 81.85 & 80.07 & 75.02 & \textbf{74.42} & \textbf{15.27} & \textbf{\textit{34.71}} \\ \midrule

\multirow{3}{*}{Gemini 1.5 Flash\cite{team2024gemini} (frames)} & - & 48.43 & 65.73 & 60.94 & 52.38 & 56.95 & 67.60 & 73.10 & 70.57 & 67.31 &6.05 & \textit{23.33} \\
 & + Sub  & 49.90 & 72.90 & 65.37 & 59.52 & 61.03 & 80.84 & 76.87 & 74.78 & 73.22 & 11.24& \textit{33.14} \\
 & + Sub \& HTP  & 48.22 & 66.36 & 62.60 & 54.76 & 57.62 & 68.91 & 73.88 & 70.49 & 67.95 &7.20 & \textit{24.40} \\
 \midrule

\multirow{3}{*}{Gemini 1.5 Pro\cite{team2024gemini} (frames)} & - & 44.44 & 65.11 & 67.31 & 54.76 & 57.20 & 68.61 & 76.30 & 73.57 & 69.28 & 8.93 & \textit{23.97} \\
 & + Sub  & 53.67 & 74.45 & 77.56 & 71.43 & 67.03 & 85.36 & 81.71 & 78.66 & 78.04 &20.75 & \textit{44.10} \\
 & + Sub \& HTP  & 54.72 & 75.39 & 78.95 & 69.05 & 68.03 & 85.96 & 81.78 & 79.95 & 78.76 &22.48 & \textit{45.16} \\ 
 \midrule

 \multirow{3}{*}{Gemini 1.5 Flash\cite{team2024gemini} (video)} & - & 37.95 & 60.44 & 53.74 & 57.14 & 49.38 & 69.21 & 70.39 & 65.48 & 63.68 &  8.65& \textit{22.62} \\
 & + Sub & 47.38 & 72.59 & 64.27 & 64.29 & 59.78 & 80.34 & 78.58 & 73.24 & \textbf{72.91} & 11.82 & \textbf{\textit{30.73}} \\
 & + Sub \& HTP   & 44.23 & 66.36 & 61.77 & 61.90 & 56.04 & 76.83 & 74.16 & 70.17 & 69.19 &\textbf{12.39} & \textit{30.23} \\
 \midrule

\multirow{3}{*}{Gemini 1.5 Pro\cite{team2024gemini} (video)} & - & 50.31 & 75.08 & 74.79 & 69.05 & 64.95 & 84.65 & 80.13 & 80.52 & 77.39 & 15.85 & \textit{38.69} \\
 & + Sub & 56.39 & 74.77 & 74.79 & 73.81 & 67.44 & 84.55 & 80.50 & 77.45 & 77.31 &\textbf{19.60} & \textbf{\textit{41.61}} \\
 & + Sub \& HTP & 53.46 & 78.19 & 75.07 & 76.19 & 67.36 & 85.26 & 82.28 & 77.77 & \textbf{78.04} &\textbf{19.60} & \textit{41.47} \\
 
 \midrule

\multirow{3}{*}{GPT-4o\tablefootnote{\url{https://openai.com/index/hello-gpt-4o/}, 2024-05-13}} & - & 60.17 & 80.37 & 79.50 & 76.19 & 71.94 & 89.17 & 85.84 & 82.54 & 82.23 & 25.07 & \textit{47.94} \\
 & + Sub & 61.01 & 80.69 & 79.78 & 76.19 & 72.44 & 89.77 & 86.05 & 82.94 & 82.64 & 25.36 & \textit{48.93} \\
 & + Sub \& HTP & 64.57 & 82.55 & 77.56 & 78.57 & 73.77 & 89.87 & 86.69 & 82.54 & \textbf{83.08} & \textbf{29.39} & \textbf{\textit{50.64}} \\
 \midrule\midrule
 Human &  & \textbf{86.67} & \textbf{90.00} & \textbf{89.19} & \textbf{100.00} & \textbf{89.09} & \textbf{92.04} & \textbf{90.14} & \textbf{92.24} & \textbf{90.85}& \textbf{41.86} & \textbf{66.20} \\
 
 \bottomrule
\end{tabular}}}
\end{table*}

We process videos according to the models' capabilities. For models that cannot process video data directly (e.g., Idefics2\cite{chen2023internvl}), we uniformly sample 16 frames from each video to serve as input. For models that accept video inputs (e.g., Gemini\cite{team2024gemini}), we use raw videos and frames respectively. Additionally, we provide video subtitles generated using Whisper \cite{radford202whisper} and incorporate them into the text prompts.

Our task is formulated as a multiple-choice VideoQA problem. Models receive the raw video or selected frames, along with a question and five options, and must select the correct option. We enforce output formatting constraints and determine selected option through exact matching, ensuring reproducible evaluation results. In \cref{sec:evaluation}, we present our specific evaluation metrics.

\subsection{Evaluation}
\label{sec:evaluation}
Our evaluation methodology comprises two components: the conventional calculation of QA accuracy across various types, and the assessment of causal chain consistency metrics. For the conventional metrics, we calculate the accuracy for each QA type as well as the overall accuracy. Furthermore, we report accuracies for the four subtypes within the MSE type, namely \textit{Emotion}, \textit{Belief}, \textit{Intent}, and \textit{Desire}.

LVLMs frequently display inconsistencies. For instance, when prompted with the question, "Why A?" a model might accurately respond, "Because of B." Yet, if we subsequently inquire, "Does B exist in the video?", the model might contradict itself by denying B's presence. Traditional accuracy metrics fail to capture this subtle yet crucial issue. To address this, we propose two new consistency metrics: \textit{\textbf{Chain Consistency}} and \textit{\textbf{Subchain Consistency}}. These metrics are designed to specifically evaluate the consistency of answers provided by LVLMs, ensuring a more reliable assessment. We denote the QA generation function is denoted as $D(*)$. $D(g^{sub})$ represents all the causality-generated QAs and all node-generated QAs from the subchain $g^{sub}$. We denote all QAs about a causal chain as $D(g)$, and $D(g)=\{D(g^{sub})|\forall g^{sub} \in g)\}$. QAs in $D(g)$ are all about the same social interaction, so it is natural use them to measure whether the entire process is fully understood.

\textit{\textbf{Chain Consistency.}}
We consider a model as having a comprehensive understanding of a social interaction process if it answers all questions correctly. The \textbf{\textit{Chain Consistency}} metric is used to meature it and calculated as:
\begin{equation}
    Cons^{c} = \frac{\sum_{g \in \mathcal{G}} \prod_{(q,a) \in G(g)} \mathbb{I} (a^{*} = a)}{| \mathcal{G} |}
    \label{eq:chain-consistency}
\end{equation}

\textit{\textbf{Subchain Consistency.}} 
To more comprehensively evaluate the social reasoning abilities of LVLMs, we also need to identify which smaller processes (i.e., subchains) the model fails to understand. We use \textbf{\textit{Subchain Consistency}} to measure the model's understanding of a subchain:
\begin{equation}
    Cons^{sc} = \frac{\sum_{g^{sub} \in \mathcal{G}^{sub}} \prod_{(q,a) \in D(g^{sub})} \mathbb{I} (a^{*} = a)}{| \mathcal{G}^{sub} |}
    \label{eq:subchain-consistency}
\end{equation}

\subsection{Generated QA Evaluation}
\label{sec:Generated QA Evaluation}
We select several LVLMs—including open-source models like Idefics2-8B and Phi-3.5, and closed-source models such as GPT4o and Gemini Pro—for testing on our benchmark, guided by models reported in previous evaluations like Video-MME and MMBench-Video. Results are presented under three settings: the basic setting (using only videos or uniformly sampled frames), ``\textbf{+Sub}'' (adding subtitles), and ``\textbf{+Sub~\&~HTP}'' (adding subtitles and a heuristic ToM prompting, which encourage models to focus on analyzing characters' mental states and event contexts to form causal chains for inference).

As shown in \cref{tab:corrected_table}, there is significant disparity in performance. Humans perform the best in all aspects, and in both reasoning accuracy and reasoning consistency. Early models like Video-LLaVA answer few chains or subchains correctly, with \( \text{Cons}^{c} \) and \( \text{Cons}^{sc} \) scores near zero. In contrast, recent models like GPT-4o achieves \( \text{Cons}^{c} = 29.39\% \) and \( \text{Cons}^{sc} = 50.64\% \). Adding subtitles generally improves performance. Incorporating heuristic ToM prompts also enhances performance, except for mPLUG-Owl3. GPT showed significant improvements with heuristic ToM prompting. Overall, GPT-4o performs the best in all models. 

\begin{figure*}[t!]
    \centering
    \begin{subfigure}{0.33\textwidth}
        \centering
        \includegraphics[width=\linewidth]{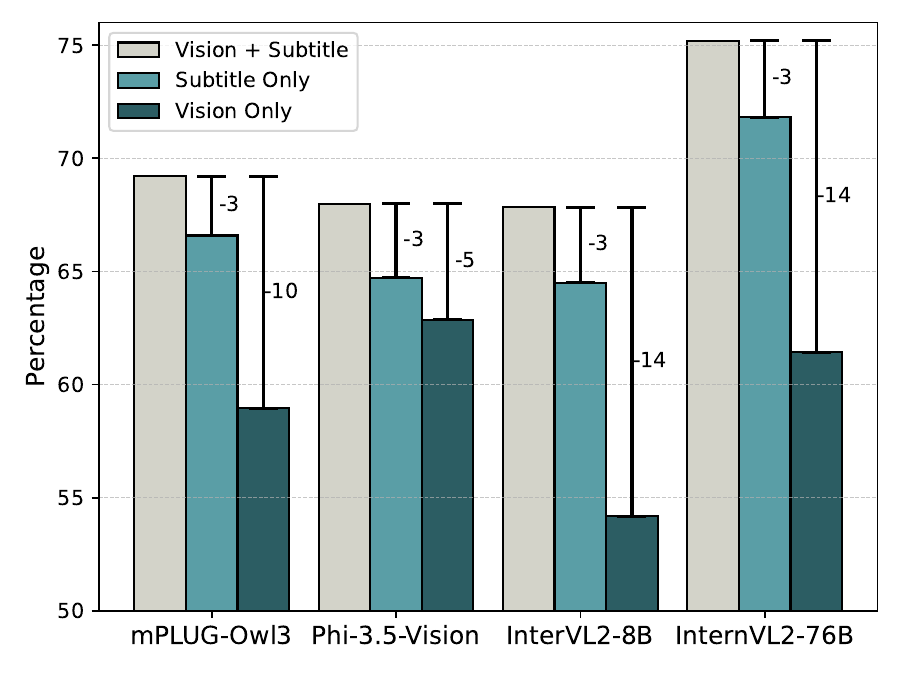}
        \caption{Overall Accuracy}
        \label{fig: modality-ablation-overall}
    \end{subfigure}
    \hfill
    \begin{subfigure}{0.33\textwidth}
        \centering
        \includegraphics[width=\linewidth]{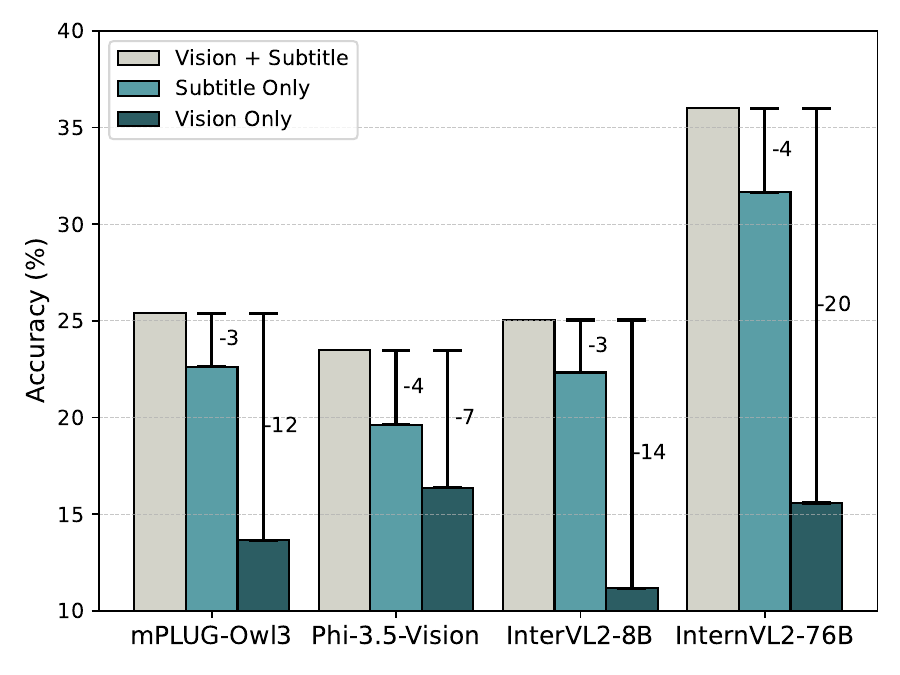}
        \caption{Subchain Consistency}
        \label{fig:modality-ablation-subchain-consistency}
    \end{subfigure}
    \hfill
    \begin{subfigure}{0.33\textwidth}
        \centering
        \includegraphics[width=\linewidth]{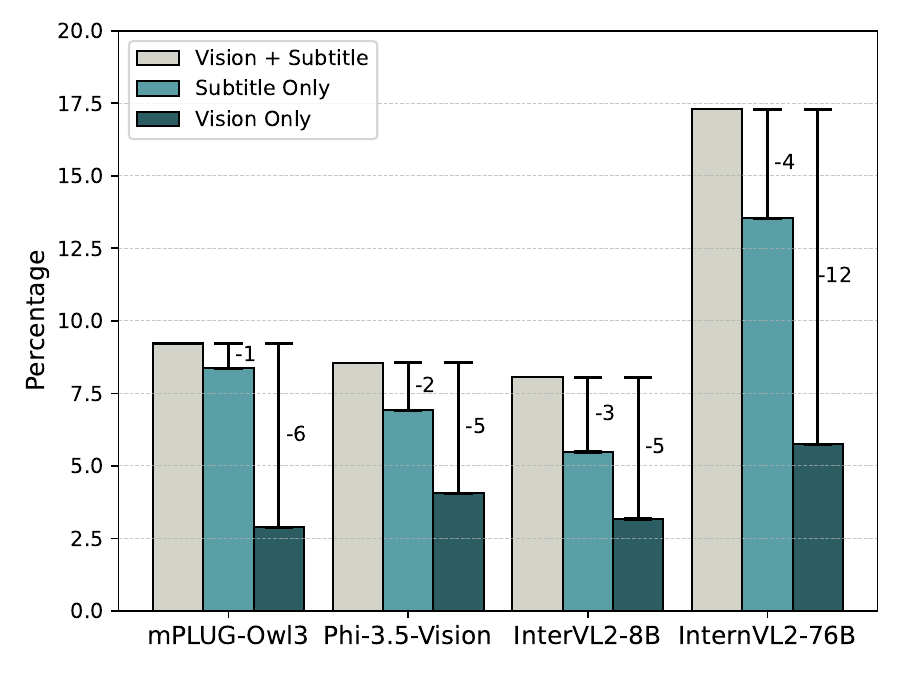}
        \caption{Chain Consistency}
        \label{fig:modality-ablation-chain-consistency}
    \end{subfigure}

    \caption{Modality ablation results of open-sourced LVLMs.}
    \label{fig:modality-ablation}
    % \vspace{-0.5cm}
\end{figure*}

\subsection{Human Designed QA Evaluation}
\label{sec:challenge-eval}

Similarly, as presented in \cref{sec:Generated QA Evaluation}, we evaluate models on human designed QAs and results are shown in \cref{tab:human_table}. Most models perform only slightly above random, with the best-performing model, GPT-4o, achieving an accuracy of merely 48.73\%, highlighting the dataset's difficulty. Human can achieve an accuracy of 80\%, which is much higher than models, indicating that SOTA LVLMs are still far from human-level social reasoning.

\begin{table}[h]
\centering
\caption{Human Designed QAs Evaluation Results (\%). Results of other LVLMs can be seen in Sec 3.2 of supp.}
\label{tab:human_table}
 \resizebox{\columnwidth}{!}{%
\begin{tabular}{llc}
\toprule
\textbf{Model} & \textbf{Method} & \textbf{Overall} \\ \midrule

Random&-& 20\\ \midrule

\multirow{2}{*}{Idefics3-8B-Llama3\cite{laurenccon2024building}} & + Sub & 24.37  \\ 
                                    & + Sub \& HTP & 23.73 \\  \midrule
                                    
\multirow{2}{*}{InternVL2-26B\cite{chen2023internvl}} & + Sub & 24.68  \\ 
                                & + Sub \& HTP & 26.27 \\  \midrule

\multirow{2}{*}{mPLUG-Owl3\cite{ye2024mplug}} & + Sub & 29.11 \\ 
                            & + Sub \& HTP & 22.79 \\  \midrule

\multirow{2}{*}{InternVL2-76B\cite{chen2023internvl}} & + Sub & 31.96 \\ 
                               & + Sub \& HTP & 29.75 \\  \midrule

\multirow{2}{*}{GPT-4o Mini\tablefootnote{\url{https://openai.com/index/gpt-4o-mini-advancing-cost-efficient-intelligence/}, 2024-07-18}} & + Sub & 30.70  \\ 
                             & + Sub \& HTP & 31.96 \\  \midrule

\multirow{2}{*}{Gemini 1.5 Pro\cite{team2024gemini} (video)} & + Sub & 34.81  \\ 
                                 & + Sub \& HTP & 39.56 \\  \midrule
\multirow{2}{*}{Gemini 1.5 Pro\cite{team2024gemini} (frames)} & +  & 39.24  \\ 
                                 & + Sub \& HTP &  38.29\\  \midrule

\multirow{2}{*}{GPT-4o\tablefootnote{\url{https://openai.com/index/hello-gpt-4o/}, 2024-05-13}} & + Sub & 48.73  \\ 
                        & + Sub \& HTP & 53.80 \\  
\midrule \midrule
Human & - & \textbf{80.06} \\                            
\bottomrule
\end{tabular}

}
\end{table}
\section{Discussion}
\label{sec:dis}

\textbf{For LVLMs, estimating mental states is more difficult than understanding events.}
Estimating mental states is an inferential task, whereas understanding events is a factual task. Inference presents inherent challenges, as it requires spatial-temporal perception, cognition, and reasoning. We must infer mental state from human's verbal and non-verbal signals. In \cref{tab:corrected_table}, the accuracies on \textit{MSE} are generally lower than on \textit{EU}. 
This significant discrepancy may explain why LVLMs struggle with social reasoning tasks.

\textbf{Models have significant shortcomings in consistent reasoning.}
InternVL2-76B, GPT-4o, Gemini 1.5 Pro, etc. all achieve over 70\% overall accuracies, but they display notably low consistency in subchains (below 51\%) and chains (below 30\%). The chain consistency and subchain consistency represent the understanding of entire and smaller social causal processes respectively. These low consistencies indicate that LVLMs struggle with fully grasping the causal relationships and dynamics in events and mental states, which suggests that they have significant deficiencies in complex social reasoning.

\textbf{Multi-step reasoning is more challenging.} To compare the abilities of single-step and multi-step reasoning, we also keep human designed QAs in \cref{subsec: data construction}. For almost all models in \cref{tab:human_table}, we can see that the overall accuracies on this data were much lower than those on generated QA-pairs compared with themselves. 
This indicates that multi-step reasoning, which is involved in human designed QAs, is more difficult than model generated single-step reasoning. It inspires us that generating multi-step causal QAs using annotated causal chains is a meaningful and valuable direction.

\textbf{Multimodal social cues are crucial for social reasoning.}
In \cref{tab:corrected_table}, nearly all models show improvements in both overall accuracy and consistencies after adding subtitles. As shown in \cref{fig:modality-ablation}, the overall accuracy and consistencies of open-sourced LVLMs drops significantly using only unimodal cues. Since language is straightforward and contains rich information, models perform better with only subtitles than with only vision. However, to achieve better performance, models still needs to utilize multimodal cues. 
Therefore, multimodal information integration and alignment is an important direction for future work.

\textbf{ToM prompting enhances social reasoning.} In our work, we use heuristic ToM prompting as a way to improve LVLMs' ToM capabilities and results show that it can enhance their social reasoning capabilities. As shown in \cref{tab:corrected_table}, ToM prompting leads to marked improvements in both accuracy and consistency across powerful LVLMs, such as GPT-4o Mini. In \cref{tab:human_table}, the positive effects of ToM prompting are more obvious. These results highlight the significant potential of ToM in boosting the social reasoning abilities of LVLMs. Although this approach is still simple, we provide a potential way to improve social reasoning capabilities.

\paragraph{Acknowledgement} This work is supported by the National Natural Science Foundation of China (62406031). We thank Miss Zhen Chen (BIGAI) for making the nice figures.

{
    \small
    \bibliographystyle{ieeenat_fullname}
    \bibliography{main}
}

\end{document}

%% file: preamble.tex
\usepackage{colortbl}

\definecolor{mygray}{gray}{.92}

\usepackage{cuted}   % 提供 \begin{strip} 环境
\usepackage{caption} % 提供 \captionof 命令，用于在非浮动体环境中添加标题
\usepackage{graphicx} % 用于插入图片
\usepackage{acronym}

\acrodef{our-dataset}[\textit{R\textsuperscript{3}\xspace-VQA}]
{\textbf{R}ead-the-\textbf{R}oom \textbf{R}easoning for \textbf{V}ideo \textbf{Q}uestion \textbf{A}nswering}

\acrodef{our-model}[\textit{R\textsuperscript{3}\xspace-Agent}]{\textbf{R}ead-the-\textbf{R}oom \textbf{R}easoning \textbf{A}gent}

\usepackage{wrapfig}
\usepackage{textcomp,scalerel}